\documentclass[smallextended]{svjour3}

\smartqed

\usepackage{amsfonts}
\usepackage{amsmath}
\usepackage{amssymb}
\usepackage{amsmath}
\usepackage[bb=dsserif]{mathalpha}
\usepackage{bm}
\usepackage{booktabs}
\usepackage{caption}
\usepackage{footnote}
\usepackage{lmodern}
\usepackage{mathtools}
\usepackage{multirow}
\usepackage{subcaption}
\usepackage{url}

\begin{document}

\title{ModelRadar: Aspect-based Forecast Evaluation}

\titlerunning{ModelRadar}

\author{Vitor~Cerqueira, Luis~Roque and Carlos~Soares}

\authorrunning{V. Cerqueira et al.}

\institute{Vitor Cerqueira \at
         Faculty of Engineering University of Porto, Porto, Portugal\\
         Laboratory for Artificial Intelligence and Computer Science (LIACC), Portugal\\
         \email{vitorc.research@gmail.com}
         \and
         Luis~Roque \at
         Faculty of Engineering University of Porto, Porto, Portugal\\
         Laboratory for Artificial Intelligence and Computer Science (LIACC), Portugal\\
         \and
         Carlos~Soares \at
         Faculty of Engineering University of Porto, Porto, Portugal\\
         Laboratory for Artificial Intelligence and Computer Science (LIACC), Portugal\\
         Fraunhofer Portugal AICOS, Portugal\\
         \and
         Corresponding author: Vitor~Cerqueira\\
         }

\date{Received: date / Accepted: date}

\maketitle

\begin{abstract}

Accurate evaluation of forecasting models is essential for ensuring reliable predictions. Current practices for evaluating and comparing forecasting models focus on summarising performance into a single score, using metrics such as SMAPE. While convenient, averaging performance over all samples dilutes relevant information about model behavior under varying conditions. This limitation is especially problematic for time series forecasting, where multiple layers of averaging--across time steps, horizons, and multiple time series in a dataset--can mask relevant performance variations.
We address this limitation by proposing ModelRadar, a framework for evaluating univariate time series forecasting models across multiple aspects, such as stationarity, presence of anomalies, or forecasting horizons.
We demonstrate the advantages of this framework by comparing 24 forecasting methods, including classical approaches and different machine learning algorithms. NHITS, a state-of-the-art neural network architecture, performs best overall but its superiority varies with forecasting conditions. For instance, concerning the forecasting horizon, we found that NHITS (and also other neural networks) only outperforms classical approaches for multi-step ahead forecasting. Another relevant insight is that classical approaches such as ETS or Theta are notably more robust in the presence of anomalies. These and other findings highlight the importance of aspect-based model evaluation for both practitioners and researchers. ModelRadar is available as a Python package.

\end{abstract}

\section{Introduction}\label{intro}

Time series forecasting is a crucial task across many domains, from supply chain management to various industrial applications. As new approaches emerge, particularly in deep learning, there is a growing need for comprehensive evaluation frameworks that go beyond simple accuracy metrics. This is especially important as forecasting models are deployed in critical real-world applications where different types of errors may have varying implications \cite{yardley2021beyond}.

The typical approach for evaluating forecasts relies on averaging performance across all samples using metrics such as SMAPE (symmetric mean absolute percentage error) \cite{makridakis2018m4}. Under this approach, a model's estimated accuracy is computed by averaging errors over multiple time steps, forecasting horizons, and across entire collections of time series.

While averaging performance into a single value provides a convenient way to compare models and select the best performer, it can mask important information about model behavior. This aggregation dilutes insights about specific conditions where relative performance differs from overall accuracy or scenarios where models behave unexpectedly. For instance, a model showing the best average performance might consistently fail in critical scenarios or underperform other alternatives under certain conditions.

The real-world applicability of a model often depends on how it performs under specific conditions\footnote{Other factors besides performance may be relevant, such as computational efficiency, ease of implementation, or interpretability, but these are out of the scope of this work.} that are not captured by averaged metrics. The practice of reporting single aggregate metrics makes it difficult to understand the robustness of forecasting models--their ability to maintain performance across varying data characteristics and forecasting conditions.

We address these limitations by proposing ModelRadar, a novel framework for evaluating univariate time series forecasting models. Our approach examines model performance across multiple aspects, such as stationarity, the presence of anomalies, or forecasting horizons. By analyzing how different models behave under specific conditions rather than relying on aggregate metrics, ModelRadar provides practitioners with detailed insights about model capabilities crucial for informed model selection and robustness evaluation.

The framework enables a systematic comparison of several forecasting methods. Our empirical study includes classical methods such as ARIMA \cite{hyndman2008automatic} and exponential smoothing \cite{hyndman2008forecasting}, machine learning regression algorithms such as LightGBM, and state-of-the-art deep learning architectures such as NHITS \cite{challu2023nhits}. While the comparison between neural networks and classical approaches has been extensively studied \cite{tang1991time,makridakis2018statistical}, they do not systematically control for different aspects relevant to forecasting problems. Our aspect-based evaluation reveals nuanced patterns in relative performance that are not captured by traditional evaluation approaches.

With ModelRadar and an extensive empirical study, we aim to address the following research questions:
\begin{itemize}
    \item \textbf{RQ1}: How can we better characterize model performance to evaluate robustness across diverse forecasting conditions and inform more reliable model selection?
    \item \textbf{RQ2}: How does the relative performance of different forecasting methods vary across different data characteristics and forecasting conditions?
    \item \textbf{RQ3}: Under what conditions do classical methods remain competitive with or outperform state-of-the-art deep learning approaches?
\end{itemize}

The results of our study indicate that NHITS \cite{challu2023nhits}, a state-of-the-art deep neural network architecture, performs best across several dimensions. However, its superiority varies with forecasting conditions. For instance, in terms of forecasting horizon, NHITS only outperforms classical approaches for multi-step ahead forecasting. When dealing with anomalies, NHITS and other machine learning approaches are outperformed by methods such as ETS or Theta. 

To promote reproducibility and enable further research in this direction, we provide both ModelRadar as a Python software package\footnote{\url{https://github.com/vcerqueira/modelradar}} and the complete experimental setup\footnote{\url{https://github.com/vcerqueira/experiments-modelradar}}.
We note that this paper extends our previous work \cite{cerqueira2024forecasting} in several important ways. First, we significantly expand the experimental study by including additional datasets and forecasting methods. Second, we formalize the ModelRadar framework, providing a systematic approach to multi-dimensional forecast evaluation. Third, we introduce new evaluation dimensions, such as stationarity and seasonality handling. These extensions provide a more comprehensive understanding of forecasting model performance across diverse conditions.

The rest of this paper is organized as follows. Section \ref{sec:background} provides background on time series forecasting, including problem definition and an overview of forecasting approaches from classical methods to recent deep learning developments. Section~\ref{sec:modelradar} presents ModelRadar, our proposed framework for aspect-based evaluation of forecasting models. Section \ref{sec:materials} describes our experimental setup, including datasets and forecasting methods. Section \ref{sec:experiments} presents our empirical results, analysing model performance across multiple dimensions. Section \ref{sec:discussion} discusses practical implications of our findings and limitations of the study. Finally, Section \ref{sec:conclusions} concludes the paper.

\section{Background}\label{sec:background}

This section overviews several topics related to our work. We start by defining the univariate time series forecasting problem within Section \ref{sec:2.1}. Then (Section~\ref{sec:2.2}), we briefly describe several forecasting methods, ranging from classical approaches to deep neural networks. We also describe previous attempts to compare different forecasting methods.
Finally, we overview evaluation practices used in forecasting problems (Section \ref{sec:2.3}).

\subsection{Time Series Forecasting}\label{sec:2.1}

A univariate time series is defined as a temporal sequence of values $Y = \{y_1, y_2, \dots,$ $y_t \}$, where $y_i \subset \mathbb{R}$ is the value of $Y$ at the $i$-th timestep and $t$ is the size of $Y$. 
We address univariate time series forecasting tasks, where the goal is to predict the value of upcoming observations of the time series, $y_{t+1}, \ldots, y_{t+H}$, where $H$ denotes the forecasting horizon. 

Forecasting problems often involve time series databases that contain multiple univariate time series. We define a time series database as $\mathcal{Y} = \{Y_1, Y_2, \dots, Y_N\}$, where $N$ is the number of time series in the collection. In these scenarios, forecasting approaches fall into one of two categories: local or global \cite{januschowski2020criteria}. Local methods build a model for each time series in a database. Classical forecasting techniques usually follow this approach. On the other hand, global methods train a single model using all time series in the database. 
Using several time series to train a model has been shown to lead to better forecasting performance \cite{godahewa2021ensembles}. The intuition for this effect is that the time series in a database are often related, for example, the demand time series of different related retail products. Global models can learn useful patterns in some time series that are not revealed in others, while local approaches only learn dependencies across time.

\subsection{Forecasting Methods}\label{sec:2.2}

There are several methods to tackle univariate time series forecasting problems. We categorize these into three main approaches: classical, machine learning regression, and deep learning. Classical approaches include well-established methods that have been foundational to the field for decades. Machine learning regression methods leverage algorithms traditionally used for supervised learning regression tasks, adapting them to time series forecasting problems. Deep neural networks represent more recent developments, with numerous architectures specifically designed to capture temporal patterns and dependencies in time series data.

\subsubsection{Classical approaches}

One of the simplest forecasting methods is naive. It predicts future values by using the last known observation. The seasonal naive method extends this principle by incorporating seasonality. Instead of using the last observation, it bases predictions on the last known value from the same seasonal period. These methods typically serve as baselines in forecasting evaluations.

ARIMA and exponential smoothing are two long-standing classical approaches to forecasting \cite{hyndman2018forecasting}. ARIMA models time series according to a linear combination of past values along with a linear combination of past errors, plus a differencing operation for integrated time series. The model order is typically selected automatically using information criteria such as AIC or BIC, which balance model complexity with goodness of fit.

Similarly to auto-regression, exponential smoothing models time series based on a linear combination of past observations. The simple exponential smoothing model involves a weighted average of the past values, where the weight decays exponentially as the observations are older \cite{gardner1985exponential}. This decay rate is controlled by a smoothing parameter that can be optimized using historical data. Several variant of exponential smoothing have been developed over the last decades that include additional components to handle different dynamics in time series.

Classical approaches typically employ a local methodology, where a separate model is fitted for each time series in a dataset. This approach allows the model parameters to be optimized specifically for the patterns and characteristics of each individual series. However, it does not leverage potential similarities or relationships between different time series in the same database, which could provide valuable information for improving forecast accuracy.

\subsubsection{Machine learning regression}

Machine learning approaches frame forecasting as a supervised learning problem using an auto-regressive modeling strategy. The idea is to transform temporal dependencies into a standard feature-target relationship using time delay embedding~\cite{bontempi2013machine}. 

Time delay embedding reconstructs a time series into the Euclidean space using sliding windows. This process creates a dataset $\mathcal{D}=\{<X_{i}, y_{i}>\}^t_{i=p+1}$ where each target $y_i$ represents the value to be predicted, and $X_i \in \mathbb{R}^p$ is its corresponding feature vector containing $p$ lags: $X_i = \{y_{i-1}, y_{i-2}, \dots, y_{i-p} \}$. This transformation allows the application of standard machine learning regression algorithms to time series data.

For global forecasting models, the training process combines data from multiple time series during the preparation stage. The training dataset $\mathcal{D}$ is created by concatenating individual time series datasets: $\mathcal{D} = \{\mathcal{D}_1, \dots, \mathcal{D}_n\}$, where $\mathcal{D}_j$ represents the embedded dataset for time series $Y_j$. The auto-regressive formulation is then applied to this combined dataset, allowing the model to learn from the entire collection of time series simultaneously, potentially improving generalization.

\subsubsection{Deep Learning}

Various types of neural network architectures have been recently developed for time series forecasting. These follow the auto-regressive formulation described above while using specialized architectures to capture more complex temporal patterns.

Architectures based on recurrent neural networks, such as the LSTM or GRU \cite{yamak2019comparison}, are more common due to their intrinsic capabilities for sequence modeling. These architectures can be coupled with temporal dilated connections to improve the modeling of long sequences \cite{chang2017dilated}.
Another notable recurrent-based architecture is DeepAR \cite{salinas2020deepar}, which uses stacked auto-regressive LSTM layers and produces probabilistic forecasts through Markov Chain Monte Carlo sampling.

Following advances in natural language processing, several methods adopted the transformer architecture, such as the Temporal Fusion Transformer \cite{lim2021temporal}, Informer \cite{zhou2021informer}, or PatchTST \cite{nie2022time}. However, recent studies have questioned the effectiveness of transformer-based approaches for forecasting tasks \cite{zeng2023transformers}.

The multi-layer perceptron (MLP), despite its simplicity, has a long history in forecasting \cite{hill1996neural}. Recent architectures such as N-BEATS \cite{oreshkin2019n} and NHITS \cite{challu2023nhits} have demonstrated that MLPs with additional structural components can achieve state-of-the-art performance. NHITS is based on stacks that contain blocks of multi-layer perceptrons (MLP) along with residual connections. The architecture behind NHITS also features other relevant aspects, such as multi-rate input sampling that models data with different scales or hierarchical interpolation for better long-horizon forecasting. NHITS has shown state-of-the-art forecasting performance relative to other deep learning approaches, including various transformers and state-of-the-art recurrent-based neural networks \cite{challu2023nhits}. DeepNPTS (Deep Non-Parametric Time Series Forecaster) is another MLP-based method that learns to sample from past observed values of time series and use these to forecast \cite{rangapuram2023deep}.

While not as widespread as recurrent or densely-connected architectures, several neural networks using convolutional layers have been developed for time series data, such as the temporal convolutional network \cite{bai2018empirical}.

Kolmogorov-Arnold Networks (KAN) \cite{liu2024kan} have recently emerged as an alternative to traditional MLPs by introducing learnable activation functions. Inspired by the Kolmogorov-Arnold Representation theorem, KANs provide theoretical guarantees for function approximation and have shown promising results in forecasting applications \cite{han2024kan4tsf}.

Deep learning models are typically trained on large collections of time series in a global fashion. This approach allows them to learn shared patterns across multiple time series while maintaining the capacity to adapt to individual characteristics. 

\subsubsection{Comparing forecasting methods}\label{sec:2.2.4}

Several previous works have compared different forecasting methods.
Hill et al. \cite{hill1996neural} pioneering work in the mid-1990s, before the so-called deep learning revolution, shows that even relatively simple MLPs exhibit a competitive performance with classical approaches such as ARIMA. Tang et al. \cite{tang1991time} also compare MLPs with ARIMA-based methods and report that MLPs have a competitive forecasting performance. One key finding is that the neural network performed better for long-term forecasting, while ARIMA was better for the short-term.

Ahmed et al. \cite{ahmed2010empirical} compare different machine learning algorithms for time series forecasting and conclude that MLPs and Gaussian Processes exhibit the best performance. 
In a seminal work, Makridakis et al. \cite{makridakis2018statistical} extend the study by Ahmed et al. \cite{ahmed2010empirical} by including classical approaches such as ARIMA or exponential smoothing. They conclude that most classical approaches, including naive, outperformed machine learning methods, including neural network algorithms. However, this study is biased towards time series dataset with a low sample size \cite{cerqueira2022case}, where neural networks become heavily over-parametrized \cite{triebe2019ar}.

The M4 forecasting competition \cite{makridakis2018m4}, which featured 100,000 from various application domains, represents an important mark for understanding the relative performance of forecasting methods. 
This competition was won by an approach called ES-RNN \cite{smyl2020hybrid} that combines exponential smoothing with an LSTM neural network trained globally.
The subsequent M5 forecasting competition \cite{makridakis2022m5} included 42,840 time series from a retail company. One of the main findings from this competition is that machine learning approaches outperformed classical methods. The winning solution was based on gradient boosting using LightGBM \cite{ke2017lightgbm}.

\subsection{Evaluating Forecasts}\label{sec:2.3}

\subsubsection{Evaluation Metrics}

There are several measures to evaluate the performance of point forecasts. These fall into different categories, such as scale-dependent, scale-independent, percentage, or relative metrics. 
Hewamalage et al. \cite{hewamalage2023forecast} survey a comprehensive list of metrics and provide recommendations for which ones should be used in different scenarios. Overall, there is no consensus concerning what the best metric is. Nonetheless, for a sufficiently large sample size, most metrics agree on what the best forecasting model is \cite{koutsandreas2022selection,cerqueira2023model}.

In the benchmark M4 forecasting competition \cite{makridakis2018m4}, two metrics were used for evaluation: SMAPE and MASE (mean absolute scaled error). These are defined as follows:
\begin{equation}
    \text{SMAPE} = \frac{100\%}{n} \sum_{i=1}^{n} \frac{|\hat{y}_i - y_i|}{(|\hat{y}_i| + |y_i|)/2}
\end{equation}

\begin{equation}
    \text{MASE} = \frac{\frac{1}{n} \sum_{i=1}^{n} | y_i - \hat{y}_i |}{\frac{1}{n-m} \sum_{i=m+1}^{n} | y_i - y_{i-m} |}
\end{equation}

\noindent where $\hat{y}_i$, and $y_i$ are the forecast and actual value for the $i$-th instance, respectively, $n$ is the number of observations and $m$ is the seasonal period.
These and other metrics are usually computed across all available predictions points, which include multiple time steps, forecasting horizons, and time series. 

\subsubsection{Open challenges}

While standard evaluation metrics provide a convenient way to compare forecasting models, the practice of averaging performance across all samples can mask important information. A single aggregate score may not reveal how models perform under different conditions or identify scenarios where their relative performance differs from the overall accuracy.

Understanding model robustness - how forecasting methods perform under challenging conditions such as anomalous periods, different forecasting horizons, or varying data characteristics is increasingly relevant for the responsible use of forecasting models. Traditional evaluation approaches that rely on averaged metrics may not adequately capture these aspects of performance. 

While other important challenges exist in forecast evaluation, such as dataset selection bias \cite{roque2024cherry}, balancing accuracy with operational constraints \cite{yardley2021beyond}, or assessing forecast uncertainty \cite{makridakis2016forecasting}, this work focuses specifically on addressing the limitations of averaged performance metrics and understanding model robustness across different conditions.

\section{ModelRadar: Aspect-Based Evaluation Framework}\label{sec:modelradar}

We propose ModelRadar, a framework for systematic evaluation of forecasting models across multiple aspects. 

\subsection{Notations}

Consider a forecasting model $M$ evaluated on a collection of time series $\mathcal{Y} = \{Y_1, ..., Y_N\}$. For each time series $Y_j$, the model produces a sequence of forecasts $\hat{Y}_j$. Let $L(Y, \hat{Y})$ denote a loss function that quantifies the forecasting error between the true values $Y$ and predictions $\hat{Y}$. We can also denote as $L_{\mathcal{Y}} = \{L(Y_j, \hat{Y}_j)\}_{j=1}^N$ the set of loss scores across the collection of time series $\mathcal{Y}$.

The standard approach for evaluating forecasting models involves reporting the average loss across all time series:
\begin{equation}
    \bar{L}_{\mathcal{Y}} = \frac{1}{N} \sum_{j=1}^N L(Y_j, \hat{Y}_j)
\end{equation}

\subsection{Performance Aggregation}

We evaluate forecasting accuracy with a metric $L$ using three complementary aggregation approaches:

\begin{itemize}
    \item Overall Performance: Following standard practice, we compute the average loss $\bar{L}_{\mathcal{Y}}$ across the entire collection of time series.
    
    \item Expected Shortfall: Adapted from financial risk analysis, the expected shortfall quantifies the expected performance on the worst-performing cases. We adopt this idea to our study and measure forecasting accuracy on the $\alpha$\% of time series where a given model shows the worst scores. Given $L_{\mathcal{Y}}$ ordered from highest to lowest value, the expected shortfall at confidence level $\alpha$ is computed as:
    \begin{equation}
        \text{ES}_{\alpha} = \frac{1}{\lfloor\alpha n\rfloor} \sum_{j=1}^{\lfloor\alpha n\rfloor} L_{\mathcal{Y}}^{(j)}
    \end{equation}
    where $L_{\mathcal{Y}}^{(j)}$ represents the j-th largest loss value. 
    
    \item Win/Loss Analysis: When comparing two models $M_1$ and $M_2$, we compute the proportion of time series where one outperforms the other:
    \begin{equation}
        \text{WR}_{1,2} = \frac{1}{n} \sum_{j=1}^n \mathbb{1}[L(Y_j, \hat{Y}_{1j}) < L(Y_j, \hat{Y}_{2j})]
    \end{equation}
    where $\mathbb{1}[\cdot]$ is the indicator function. Win/loss ratios provide a non-parametric way of comparing models that mitigate the effect of outliers.
\end{itemize}

\subsection{Evaluation Conditions}

To provide a more granular analysis of model performance, we evaluate forecasting accuracy under different conditions. 
Given a condition $c$, we can analyze model performance either at the series level (e.g., seasonal strength) or observation level (e.g., presence of anomalies). For series-level conditions, we denote the loss on the subset of time series satisfying $c$ as:
\begin{equation}
    \bar{L}_{\mathcal{Y}}^c = \frac{1}{|\mathcal{Y}^c|} \sum_{Y_j \in \mathcal{Y}^c} L(Y_j, \hat{Y}_j)
\end{equation}
where $\mathcal{Y}^c \subseteq \mathcal{Y}$ represents time series satisfying condition $c$.

For observation-level conditions, we denote the loss on individual observations satisfying $c$ as:
\begin{equation}
    \bar{L}_{\mathcal{Y}}^c = \frac{1}{\sum_{j=1}^n |Y_j^c|} \sum_{j=1}^n \sum_{y_i \in Y_j^c} L(y_i, \hat{y}_i)
\end{equation}
where $Y_j^c$ represents the subset of observations in time series $j$ that satisfy condition $c$.

In this work, we analyze univariate time series forecasting accuracy across various data and problem characteristics. The analyzed conditions provide a framework that can be extended to incorporate additional dimensions of analysis.

\subsubsection{Data characteristics}

Data characteristics represent inherent properties of the time series:

\begin{itemize}

    \item Stationarity: We use the Kwiatkowski-Phillips-Schmidt-Shin (KPSS) test \cite{kwiatkowski1992testing} to assess stationarity. The test evaluates the null hypothesis that a time series is level stationary, against the alternative that it is non-stationary due to the presence of a unit root. 
   
   \item Seasonality: We apply the seasonal strength test proposed by Wang et al. \cite{wang2006characteristic} to detect whether a time series exhibits significant seasonality. 
    
    \item Sampling Frequency: Performance is analyzed across different temporal granularities, such as monthly and quarterly, to understand how models adapt to varying sampling frequencies.

    \item Anomalous Observations: We analyse performance on anomalous observations, which helps understand model robustness to unexpected observations. In this work, we define a time series anomaly as observations falling outside the 99\% prediction interval of the seasonal naive model.
\end{itemize}

\subsubsection{Problem Characteristics}

We also analyze performance according to characteristics of the forecasting task:

\begin{itemize}
    \item Forecasting Horizon: We evaluate models at different prediction horizons to understand how their relative performance varies between short-term and long-term forecasting.
    
    \item Problem Difficulty: Some time series exhibit patterns that are easier to model than others. We quantify series difficulty based on the overall forecasting accuracy of the seasonal naive baseline, allowing us to assess how model performance varies with problem hardness. Specifically, a time series is considered a hard problem if its seasonal naive loss score falls in the highest 10\% of the loss distribution across all series.

\end{itemize}

\section{Materials and Methods}\label{sec:materials}

This section describes the experimental setup used to demonstrate the ModelRadar evaluation framework. First, we present the datasets and summarize their characteristics (Section \ref{sec:data}). Then, we describe the forecasting methods included in our analysis, including classical approaches and machine learning methods (Section \ref{sec:methods}). Finally, we detail our evaluation protocol and implementation of the framework components (Section \ref{sec:perfestimation}).

\subsection{Datasets}\label{sec:data}

We use the following benchmark datasets that were part in past forecasting competitions: M1 \cite{makridakis1982accuracy}, M3 \cite{makridakis2000m3}, M4 \cite{makridakis2018m4}, and Tourism \cite{athanasopoulos2011tourism}. These datasets contain time series with varying sampling frequencies. For conciseness, we focus on time series that exhibit either a monthly or quarterly frequency.
Table \ref{tab:data} provides a summary of the datasets. Our analysis encompasses 75,797 time series with over 14 million observations in total. 

\begin{table}
\caption{Summary of the datasets: number of time series and number of observations}
\label{tab:data}
\begin{tabular}{llr@{\hskip 0.3cm}r}
\toprule
 & Frequency & \# time series & \# observations \\
\midrule
\multirow[t]{2}{*}{M1} & Monthly & 617 & 44892 \\
 & Quarterly & 203 & 8320 \\
\multirow[t]{2}{*}{M3} & Monthly & 1428 & 167562 \\
 & Quarterly & 756 & 37004 \\
\multirow[t]{2}{*}{M4} & Monthly & 48000 & 11246411 \\
 & Quarterly & 24000 & 2406108 \\
\multirow[t]{2}{*}{Tourism} & Monthly & 366 & 109280 \\
 & Quarterly & 427 & 42544 \\
\midrule
Total &  & 75797 & 14062121 \\
\bottomrule
\end{tabular}
\end{table}

In terms of input size\footnote{also referred to as the number of lags, or lookback window}, we use 24 lags for monthly series and 8 lags for quarterly series. The forecasting horizon is set to 12 steps for monthly series and 4 steps for quarterly series, including one complete seasonal cycle in both input and forecast periods.

\subsection{Forecasting Models and Training Protocol}\label{sec:methods}

This section describes the forecasting methods included in our experimental study. Overall, we include a total of 24 approaches, including 8 classical forecasting techniques, 4 machine learning regression methods, and 12 deep neural networks. This diverse set of approaches allows us to evaluate our framework across different modeling paradigms.
The following list describes the classical approaches:

\begin{itemize}
    \item \texttt{AutoARIMA} \cite{hyndman2008automatic}: The auto-regressive integrated moving average method, optimized using the Akaike Information Criterion (AIC). This method has been a standard benchmark in univariate time series forecasting problems. In the interest of computational efficiency, we restrict the model hyperparameter optimization to 20 iterations.

    \item \texttt{AutoETS} \cite{hyndman2008forecasting}: The error, trend, and seasonality exponential smoothing method, also optimized using AIC. The method automatically selects the most appropriate type of exponential smoothing based on the input data.

    \item \texttt{SeasonalNaive}: A baseline method that sets forecasts to the last known observation of the same seasonal period. 

    \item \texttt{RWD} (Random walk with drift) \cite{hyndman2018forecasting}: a variant of the naive method where the forecasts are adjusted according to the historical average of the time series;

    \item \texttt{SESOpt} \cite{hyndman2008forecasting}: Simple exponential smoothing with the smoothing parameter optimized by squared error minimization.

    \item \texttt{AutoTheta} \cite{assimakopoulos2000theta}: The Theta method, which combines exponential smoothing with a term that models the long-term trend component of the time series.

\end{itemize}

\noindent All classical approaches follow a local training methodology, with a model being fitted for each time series in a given collection. 

Standard machine learning regression algorithms and deep neural networks follow a global training approach. For each method, a model is trained for each dataset listed in Table \ref{tab:data}. For instance, one model is created with all monthly time series in the M3 dataset.
While classical methods with their local approach have shown robust performance over the years, the global training paradigm offers potential advantages by learning from multiple time series simultaneously.

The four machine learning regression algorithms are the following:
\begin{itemize}
    \item AutoLasso: Linear regression with LASSO (L1) regularization
    \item AutoRidge: Linear regression with ridge (L2) regularization
    \item AutoLightGBM: LightGBM, a gradient boosting framework using decision trees 
    \item AutoXGBoost: XGBoost, another implementation for gradient boosted trees
\end{itemize}

The 12 neural network architectures are the following:

\begin{itemize}
    \item AutoNHITS: Neural Hierarchical Interpolation for Time Series, featuring multi-rate sampling and hierarchical structure
    \item AutoDeepAR, AutoDeepAR-median: Autoregressive RNN for probabilistic forecasting, with mean and median prediction variants. We remark that while DeepAR follows a probabilistic optimization process, our study is focused on point forecasting.
    \item AutoKAN: a Kolmogorov-Arnold Network that uses learnable spline functions as its approximators
    \item AutoPatchTST: a Transformer-based architecture that segments time series into patches as input tokens for efficient computation
    \item AutoTFT: Temporal Fusion Transformer, a Transformer-based architecture that combines recurrent layers for local processing and interpretable self-attention layers for long-term dependencies
    \item AutoGRU: Gated Recurrent Unit network for sequence modeling
    \item AutoLSTM: Long Short-Term Memory network with enhanced gradient flow
    \item AutoDilatedRNN: Recurrent network with dilated temporal connections. The type of recurrent layer (e.g. GRU, LSTM) is selected during the optimization process.
    \item AutoMLP: Multi-Layer Perceptron with fully connected layers. The number of layers and hidden units is selected during the optimization process.
    \item AutoDeepNPTS: The Deep Non-Parametric Time Series Forecaster is an MLP-based method that learns to sample from past observed values of time series and use this to forecast. 
    \item AutoDLinear: DLinear combines simple linear forecasting by separately modeling trend and seasonal components through decomposition and dedicated linear layers.
    \item AutoTCN: a Temporal Convolutional Network with dilated causal convolutions.
\end{itemize}

We resorted to the nixtla framework\footnote{\url{https://nixtlaverse.nixtla.io/}} to implement and optimize all the above methods, specifically the \textit{statsforecast}, \textit{mlforecast}, and \textit{neuralforecast} Python packages. 
The name of each method represents the name of the corresponding implementation.

For all algorithms based on machine learning, including neural networks, we perform hyperparameter optimization using random search. From a predefined pool of possible configurations, we randomly sample and evaluate 20 configurations for each method using a validation set (detailed in the next section). The optimization process includes both model-specific parameters and basic data preprocessing choices (normalization strategy: none, standard scaling, or robust scaling) without explicitly modeling trend or seasonality unless part of the method's design.  The best-performing configuration is then used to retrain the model on the complete training data. We use the configuration pool available in the corresponding implementation. 

\subsection{Evaluation Protocol}\label{sec:perfestimation}

We use SMAPE (defined in Section \ref{sec:2.3}) as the evaluation metric, given its widespread use in forecasting competitions and its ability to handle different scales. For the expected shortfall analysis, we set $\alpha$ to 10\%, meaning this metric measures performance in the worst 10\% of time series for a given model.

For each time series, we hold out the last H observations for testing, where H represents one complete forecasting horizon (12 steps for monthly series and 4 steps for quarterly series). The remaining observations form the training set. During the hyperparameter optimization phase, we apply the same principle: the last H observations of the training set are used for validation, while earlier observations are used for model training.

While models are trained separately for each dataset and frequency combination listed in Table \ref{tab:data}, we conduct our evaluation framework analysis on the complete collection of time series. This approach provides a comprehensive assessment of model performance across different data sources and characteristics.

\section{Experiments}\label{sec:experiments}

We apply our aspect-based evaluation framework to compare the 24 forecasting methods described in Section \ref{sec:methods}. First, we present a preliminary analysis that contains all 24 models (Section \ref{sec:preanalysis}). A subset of these models will then be evaluated with ModelRadar. We present overall performance metrics across all time series (Section \ref{sec:overall}). Then, we analyze how model performance varies according to different data characteristics (Section \ref{sec:data_char}), including trend strength, seasonal strength, anomaly status, and sampling frequency. Finally, we examine performance across two different problem characteristics (Section \ref{sec:prob_char}), namely the forecasting horizon and the problem difficulty. Throughout our analysis, we use color coding to distinguish between different types of methods: classical approaches are shown in brown, machine learning regression methods in turquoise, and neural networks in purple.

\subsection{Preliminary analysis}\label{sec:preanalysis}

Before applying our complete aspect-based evaluation framework, we present results for all 24 models across different evaluation dimensions. While these initial results provide a broad overview of model performance, our subsequent detailed analysis will focus on a subset of models - specifically, the top 3 performers across each dimension. This approach ensures we capture the most relevant methods while keeping the analysis focused and concise.

\begin{figure}[!ht]
    \centering
    \includegraphics[width=\textwidth, trim=0cm 0cm 0cm 0cm, clip=TRUE]{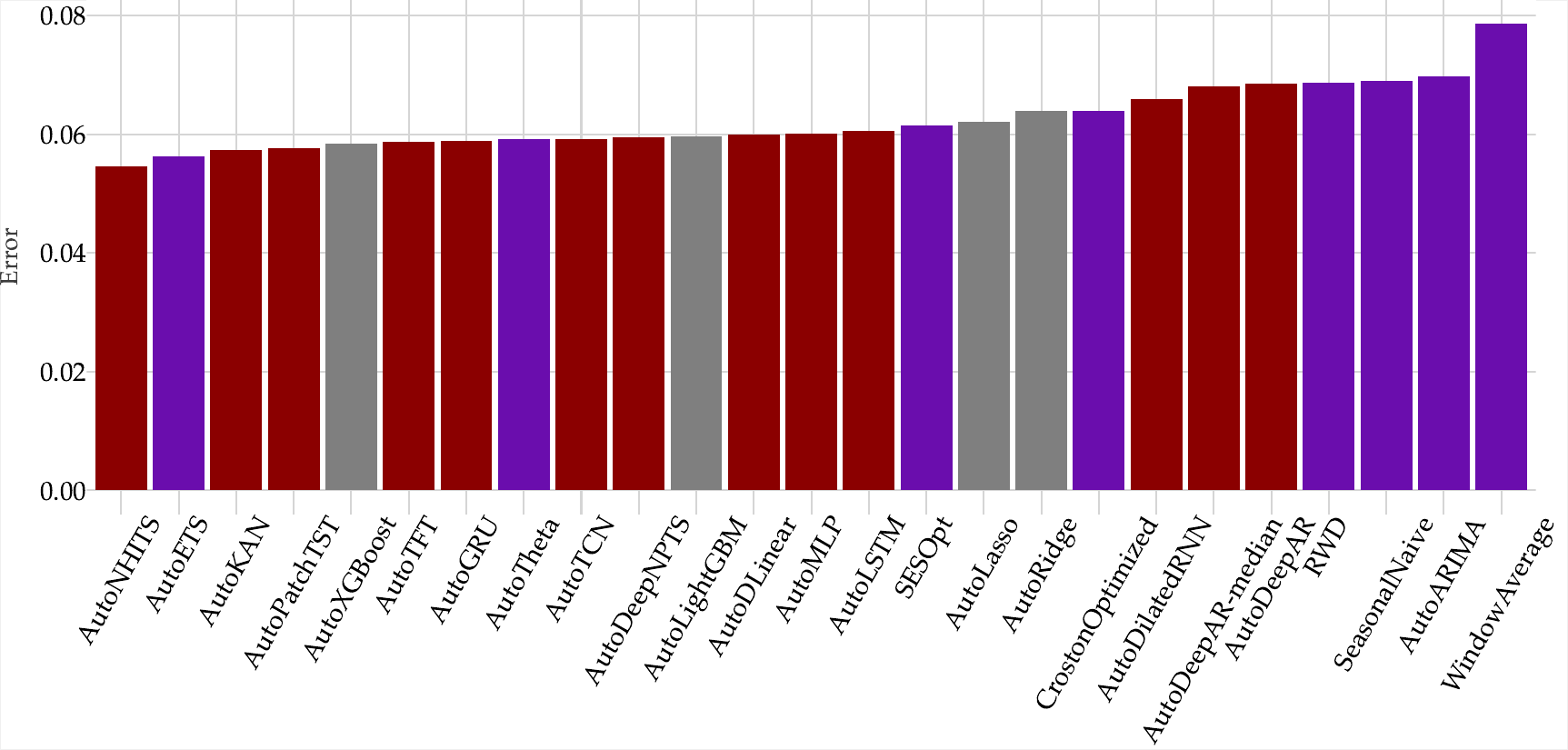}
    \caption{SMAPE scores of each model across all datasets. }
    \label{fig:plot1_preanalysis}
\end{figure}

Figure~\ref{fig:plot1_preanalysis} presents the SMAPE scores for all models across all datasets. The results show that \texttt{AutoNHITS} achieves the lowest error, followed closely by \texttt{AutoETS} and \texttt{AutoKAN}. Notably, most models achieve SMAPE scores between 0.05 and 0.07, with some classical methods (shown in brown) generally performing competitively with other approaches. The \texttt{WindowAverage} method shows the highest error, suggesting that simple averaging strategies are insufficient for univariate time series forecasting tasks.

\begin{figure}[!ht]
    \centering
    \includegraphics[width=.7\textwidth, trim=1.5cm 4.5cm 1.5cm 3cm, clip=TRUE]{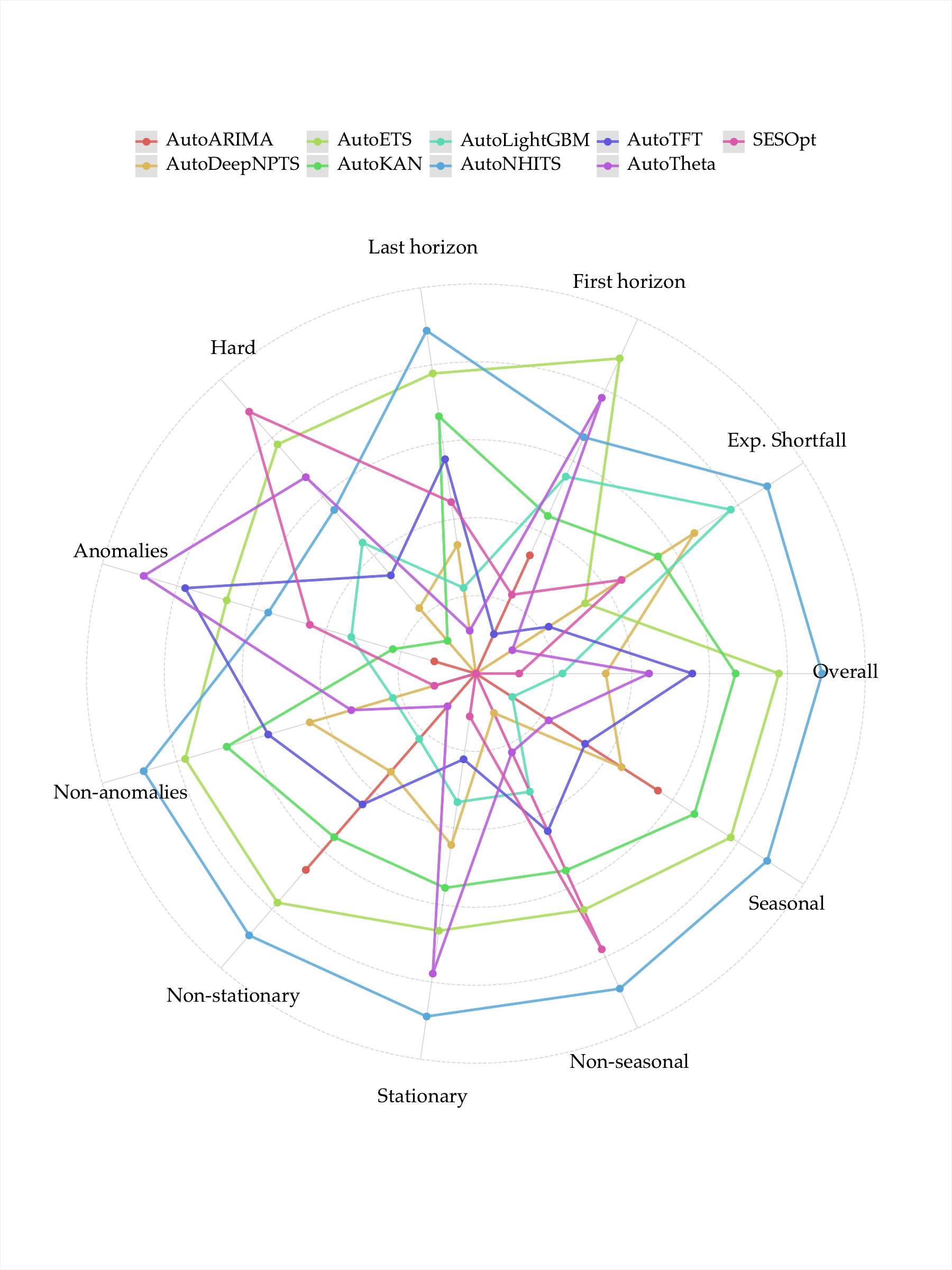}
    \caption{Rank of each model across different dimensions. Values far from the center represent better rank.}
    \label{fig:plot16_all_spider}
\end{figure}

Figure~\ref{fig:plot16_all_spider} shows a radar plot illustrating the rank of each model across several different dimensions: overall performance, expected shortfall, stationarity, seasonality, anomalies, problem difficulty, and horizon. Values far from the center represent better rank. This visualization encapsulates the results of ModelRadar and reveals several interesting patterns:
\begin{itemize}
    \item \texttt{AutoNHITS} shows the best performance across most dimensions, including overall accuracy or expected shortfall;
    \item \texttt{AutoETS} demonstrates balanced performance, with a particularly strong result in the first forecast horizon but a relatively poor one on expected shortfall;
    \item Classical methods such as \texttt{AutoTheta} perform notably well on series with anomalies, suggesting their robustness to unexpected patterns
    \item Some models show highly variable relative performance - strong in some dimensions but weak in others - highlighting the importance of aspect-based evaluation
\end{itemize}

Based on this comprehensive analysis, we identify 9 models that rank among the top 3 performers across at least one dimension: \texttt{AutoARIMA}, \texttt{AutoETS}, \texttt{AutoLightGBM}, \texttt{AutoTFT}, \texttt{SESOpt}, \texttt{AutoDeepNPTS}, \texttt{AutoKAN}, \texttt{AutoNHITS}, and \texttt{AutoTheta}. These models, along with \texttt{SeasonalNaive} that provides a baseline for forecasting accuracy, form the basis for our detailed ModelRadar analysis in subsequent sections. 

\subsection{Overall performance}\label{sec:overall}

\begin{figure}[!t]%
    \centering
    \subfloat[\centering Average SMAPE]{{\includegraphics[width=.48\textwidth]{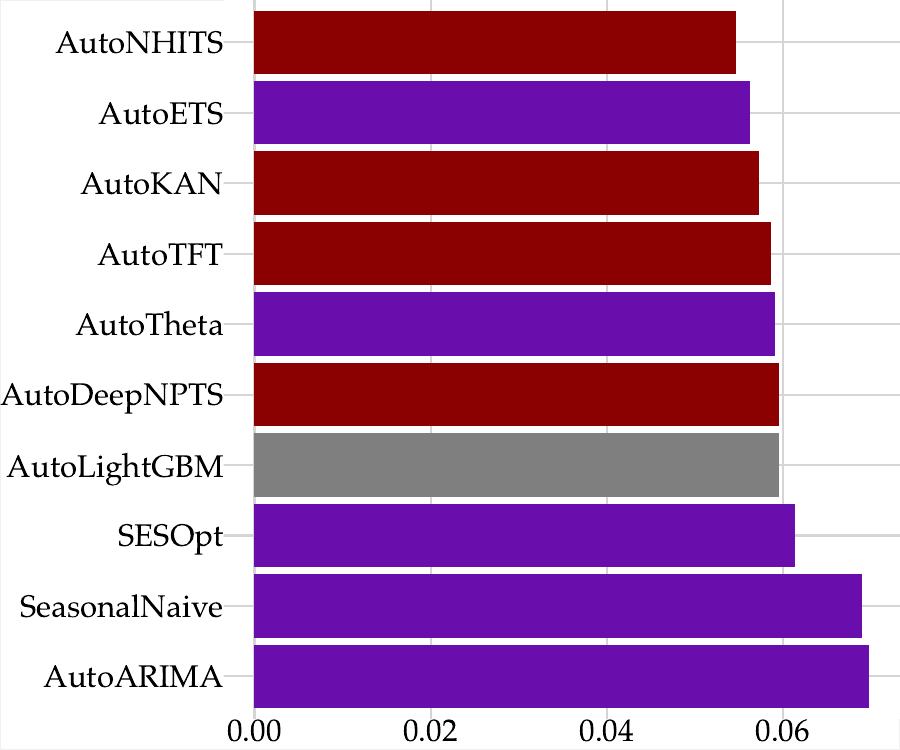} }}%
    \quad
    \subfloat[\centering SMAPE expected shortfall]{{\includegraphics[width=.48\textwidth]{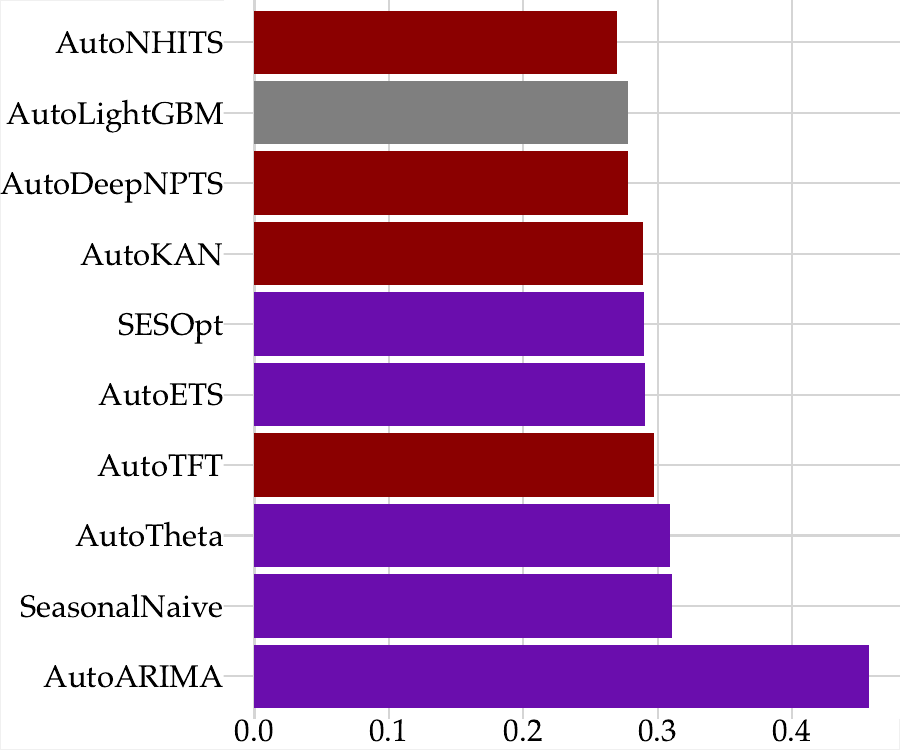}}}%
    \caption{Average SMAPE (a) and expected shortfall (b) for each model across all time series}%
    \label{fig:overall_smape}%
\end{figure}

We start applying the aspect-based evaluation framework by summarizing forecasting accuracy across all time series using SMAPE. The results are shown in Figure~\ref{fig:overall_smape}a. \texttt{AutoNHITS} presents the best average performance, followed by \texttt{AutoETS} and \texttt{AutoKAN}. Among classical methods, both \texttt{AutoETS} and \texttt{AutoTheta} show competitive performance. On the other hand, \texttt{AutoARIMA} exhibits the highest error. We recall that the configuration search process of ARIMA was restricted to 20 iterations to make it computationally feasible.

Figure~\ref{fig:overall_smape}b shows the SMAPE expected shortfall (c.f. Section~\ref{sec:perfestimation}), which assesses model performance in their worst cases. Here, the ranking of methods differs notably from the average performance. \texttt{AutoNHITS} maintains strong performance, but some methods such as \texttt{AutoETS} or \texttt{AutoTFT} struggle in terms of worst-case scenarios. Notably, while \texttt{AutoLighGBM} shows a poor rank in overall SMAPE, its expected shortfall is topped only by \texttt{AutoNHITS}. 

\begin{figure}[!t]%
    \centering
    \subfloat[\centering ROPE=0\%]{{\includegraphics[width=.48\textwidth]{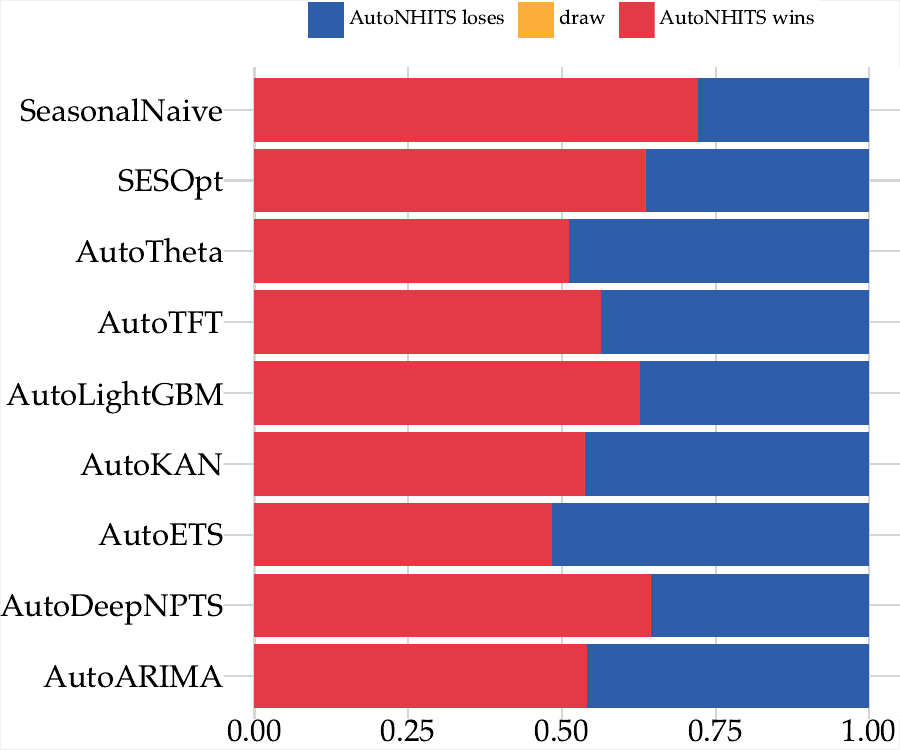} }}%
    \quad
    \subfloat[\centering ROPE=10\%]{{\includegraphics[width=.48\textwidth]{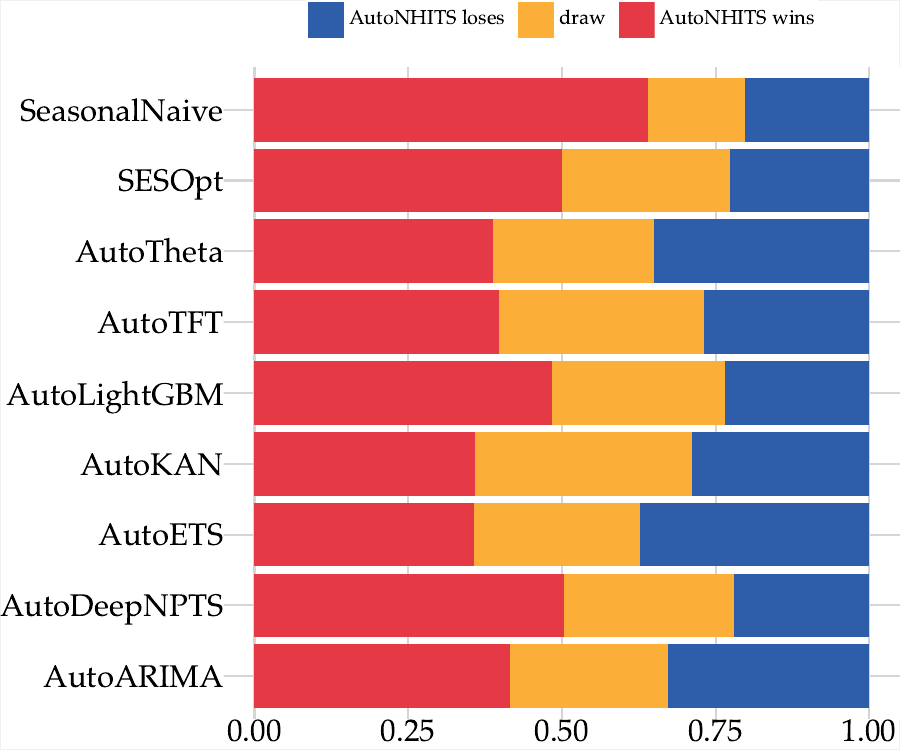}}}%
    \caption{Probability of AutoNHITS outperforming other approaches across all time series}%
    \label{fig:win_loss}%
\end{figure}

We also conducted a win/draw/loss analysis using \texttt{AutoNHITS} as reference. Figure~\ref{fig:win_loss} shows that, while \texttt{AutoNHITS} exhibits the overall best performance, there is a reasonable chance that it is outperformed by any other method. To account for small performance differences, we also analyzed the results using practical equivalence \cite{kruschke2018rejecting}. We set the region of practical equivalence (ROPE) to 10\%, considering two models to perform similarly if their absolute percentage difference in SMAPE is below this threshold. The results (Figure~\ref{fig:win_loss}b) reveal a substantial proportion of cases where methods perform equivalently, particularly classical approaches such as \texttt{AutoETS}. This suggests that while \texttt{AutoNHITS} leads in average performance, its advantage may not be practically significant in many cases. In the case of \texttt{AutoETS}, its result is consistent to the expected shortfall scores. More precisely, \texttt{AutoETS} shows comparable win/draw/loss scores with \texttt{AutoNHITS}, but poorer expected shortfall--meaning that it's worst case scenarios are worse than \texttt{AutoNHITS}', consequently leading to a poorer average performance.

These initial results motivate a deeper analysis of when and why different methods excel or struggle, which we explore through various data characteristics in the following section.

\subsection{Performance by data characteristics}\label{sec:data_char}

\subsubsection{Sampling frequency}

\begin{figure}[!ht]
    \centering
    \includegraphics[width=.95\textwidth, trim=0cm 0cm 0cm 0cm, clip=TRUE]{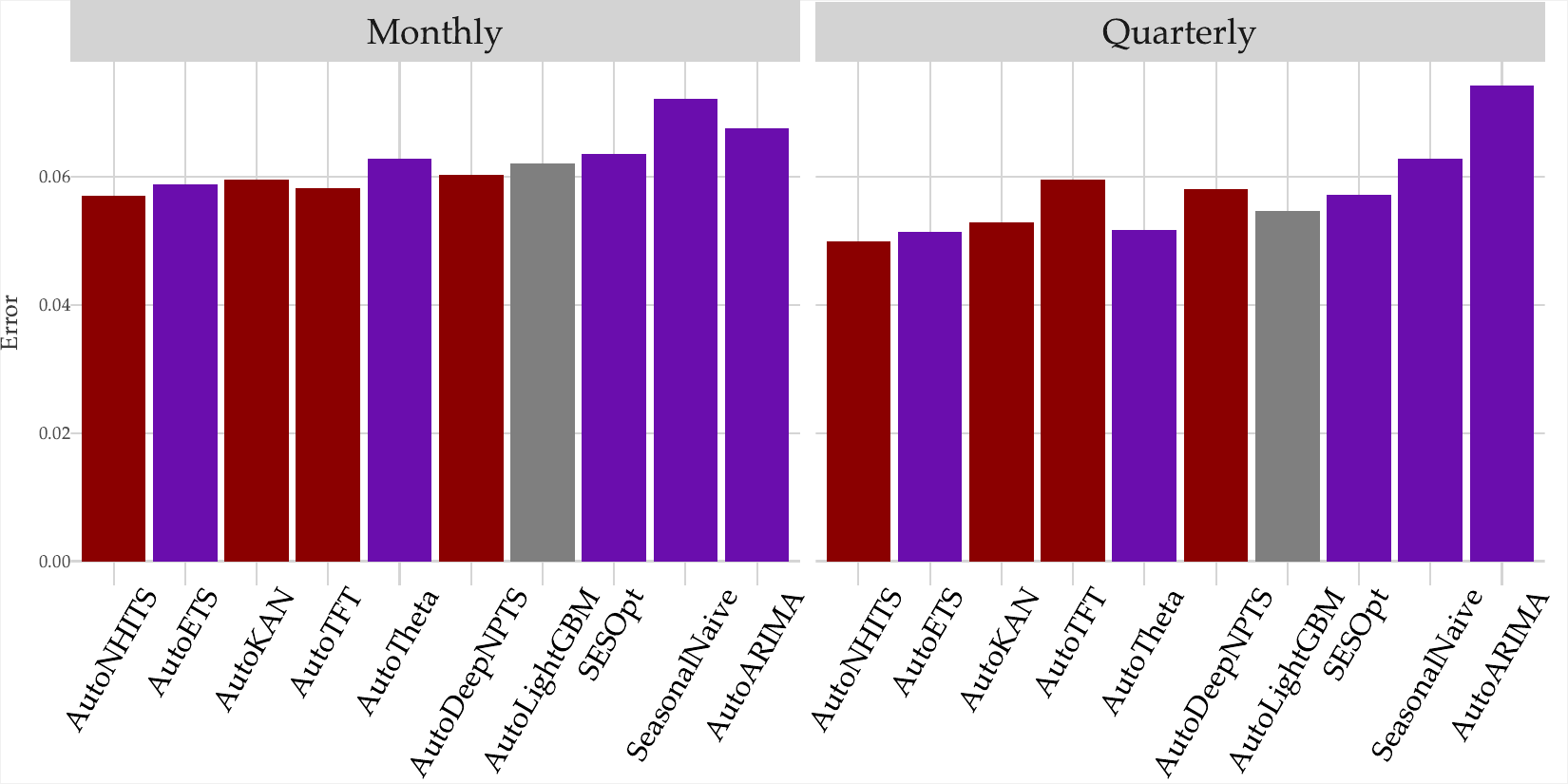}
    \caption{SMAPE scores of each model across each sampling frequency.}
    \label{fig:plot11_freq}
\end{figure}

Figure \ref{fig:plot11_freq} shows the SMAPE scores controlling for sampling frequency. The relative performance of models varies considerably between monthly and quarterly data. \texttt{AutoNHITS} maintains the leading position in both frequencies, but the competitors for second place change considerably.
For monthly data, transformer-based \texttt{AutoTFT} ranks second. However, it shows notably worse relative performance on quarterly data - \texttt{AutoTFT} drops to third-last position before \texttt{SeasonalNaive} and \texttt{AutoARIMA}, while \texttt{AutoDeepNPTS} also shows notable rank deterioration.

\texttt{AutoETS} shows strong performance across both frequencies - maintaining competitive performance in monthly data and ranking second in quarterly data. \texttt{AutoTheta} shows an interesting pattern: while showing moderate performance on monthly data, it becomes highly competitive for quarterly series, ranking third. The performance shifts across frequencies highlight the importance of sampling frequency in model selection, though further research with additional frequencies and comparable sample sizes would be needed to draw broader conclusions about the effect of data granularity on forecasting accuracy.

\subsubsection{Stationarity}

\begin{figure}[!ht]
    \centering
    \includegraphics[width=\textwidth, trim=0cm 0cm 0cm 0cm, clip=TRUE]{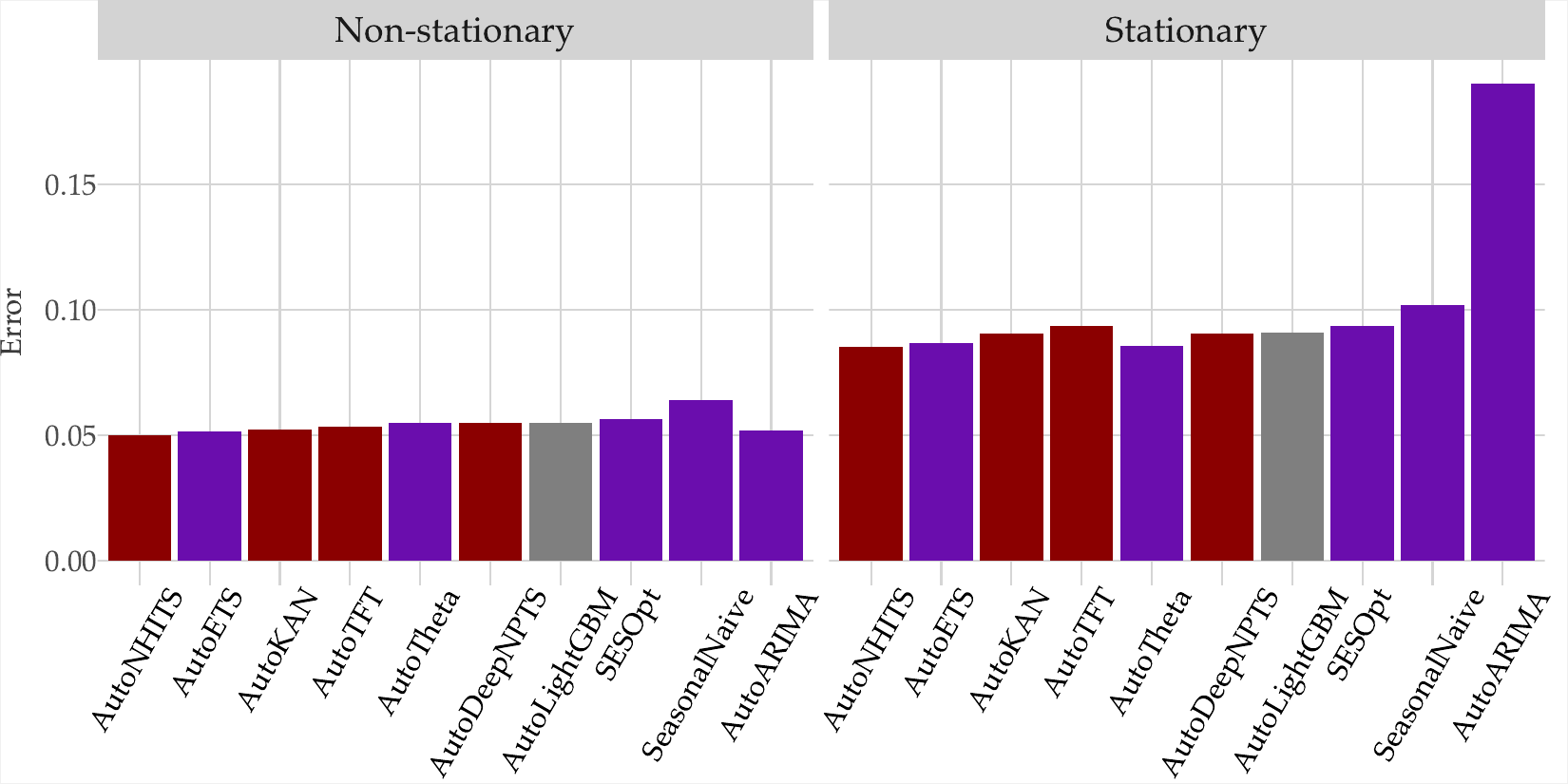}
    \caption{SMAPE scores of each model controlling for stationarity condition of time series.}
    \label{fig:plot13_trend}
\end{figure}

Figure \ref{fig:plot13_trend} reveals interesting shifts in relative performance between series that are level stationary or not according to the KPSS test. While \texttt{AutoNHITS} shows superior performance on non-stationaty time series, \texttt{AutoTheta} emerges as the best performer on stationary ones. \texttt{AutoTFT} exhibits a notable decrease in rank on stationary time series. Conversely, \texttt{AutoARIMA} is particularly well-suited for non-stationary time series. This may be explained in part due to the reduction of the configuration search space used in our experiments (c.f. Section \ref{sec:methods}).

\subsubsection{Seasonality handling}

\begin{figure}[!ht]
    \centering
    \includegraphics[width=\textwidth, trim=0cm 0cm 0cm 0cm, clip=TRUE]{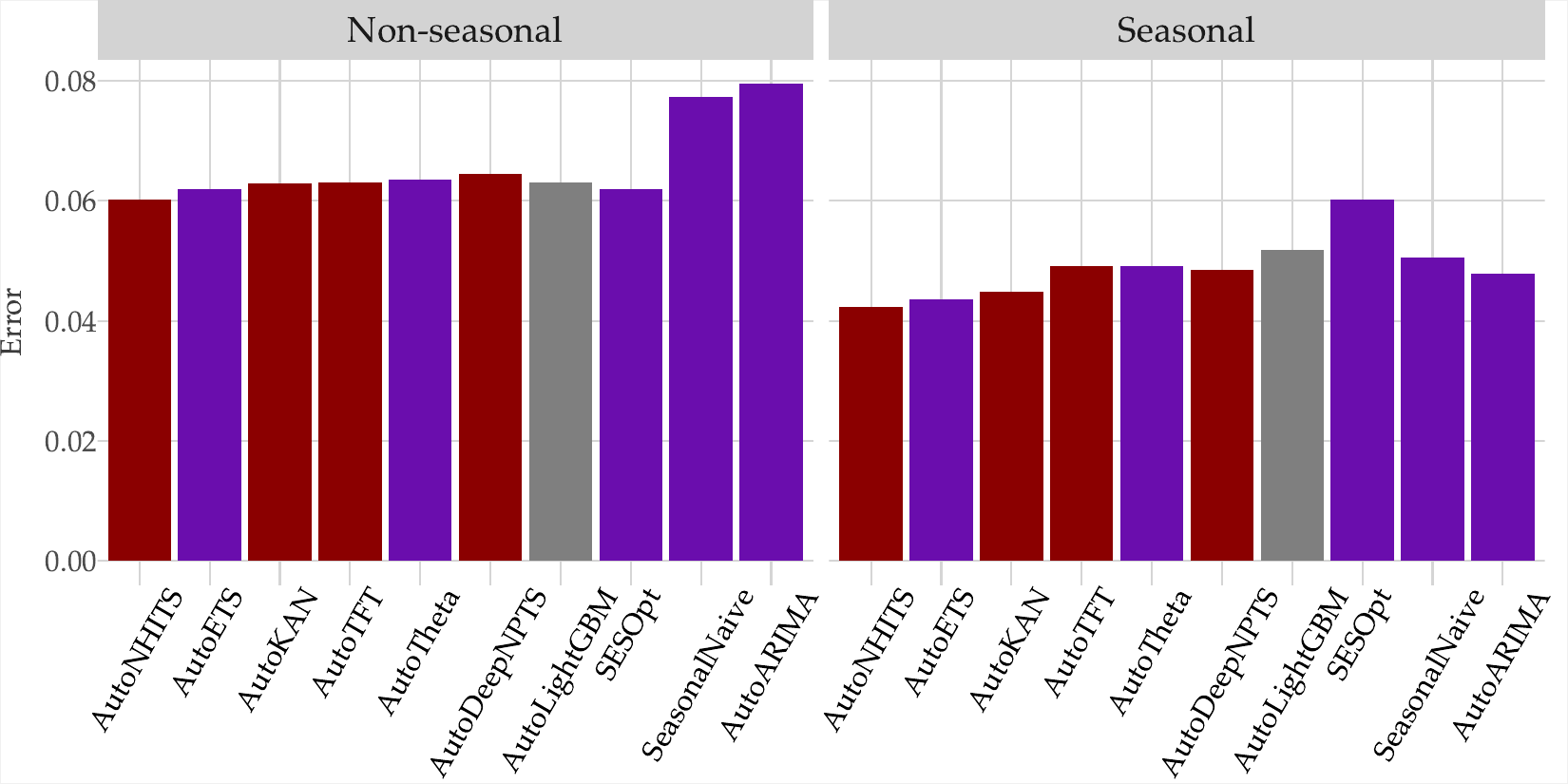}
    \caption{SMAPE scores of each model controlling for seasonality condition of time series.}
    \label{fig:plot14_seas}
\end{figure}

Figure~\ref{fig:plot14_seas} shows the forecasting accuracy controlling for seasonality status. The results reveal interesting patterns in how different methods handle seasonal patterns.

\texttt{AutoNHITS} and \texttt{AutoETS} demonstrate robust performance across both conditions, ranking first and second, respectively. This suggests their ability to effectively model time series regardless of the presence of seasonal patterns.
Methods show contrasting abilities to handle seasonality. While \texttt{SESOpt} performs competitively on non-seasonal data, it shows the highest error rates when seasonality is present. Conversely, \texttt{SeasonalNaive} and \texttt{AutoARIMA} are particularly effective on seasonal time series but struggle with non-seasonal ones - an expected behavior for \texttt{SeasonalNaive} given its design for seasonal forecasting.

\subsubsection{Anomaly handling}

Time series often exhibit unexpected or anomalous observations. These instances can significantly impact the corresponding application domain, making it important to accurately forecast such cases.

\begin{figure}[!ht]
    \centering
    \includegraphics[width=\textwidth, trim=0cm 0cm 0cm 0cm, clip=TRUE]{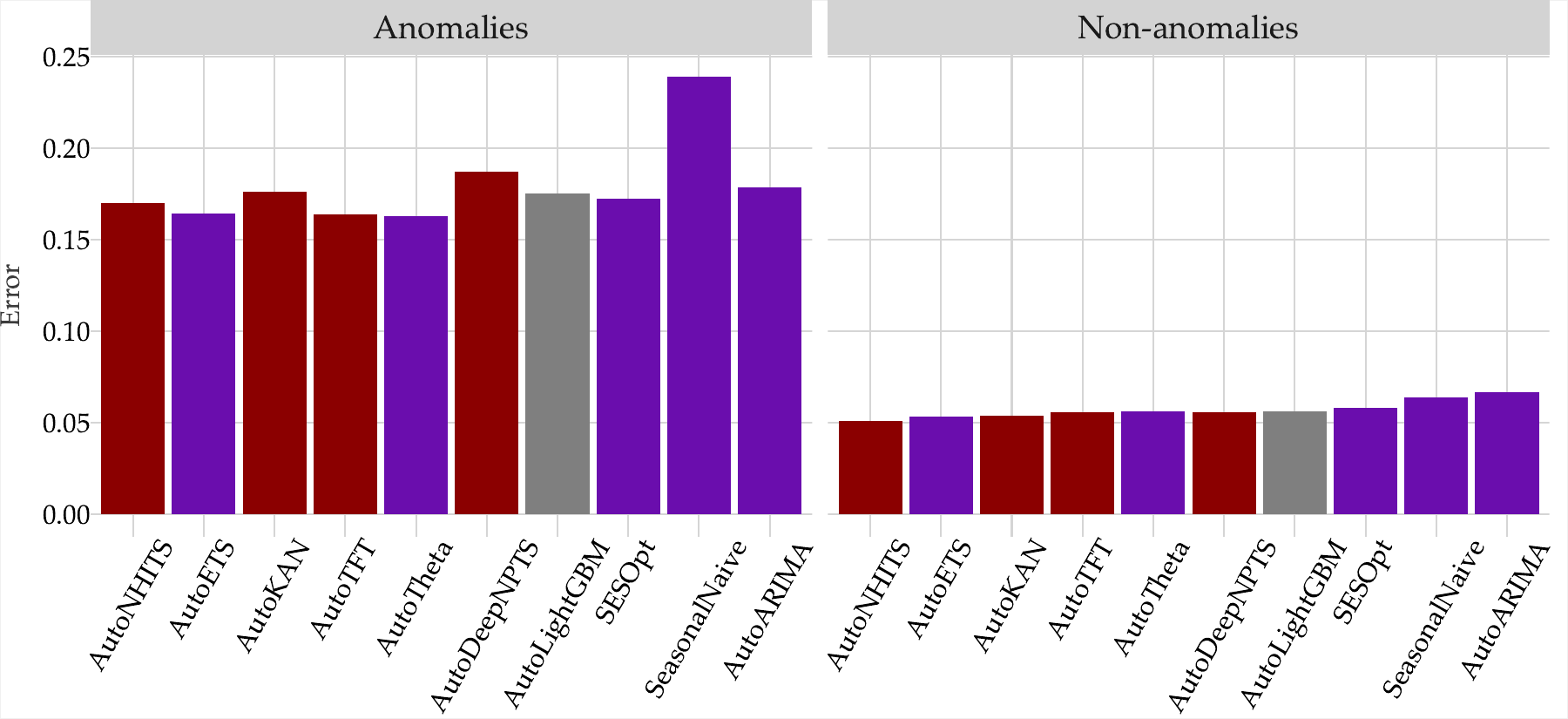}
    \caption{SMAPE scores of each model on normal and anomalous observations.}
    \label{fig:plot7_anomaly}
\end{figure}

Figure \ref{fig:plot7_anomaly} shows the performance of each model controlling for whether observations are anomalous. As expected, error scores are significantly higher when forecasting anomalous observations.

The relative performance of models varies considerably between conditions. While \texttt{AutoNHITS} demonstrates superior performance on normal observations, it shows less robustness to anomalies, where several other approaches perform better. Notably, \texttt{AutoETS}, \texttt{AutoTFT}, and \texttt{AutoTheta} handle anomalous observations more effectively. 
These results highlight that model selection might need to consider the presence and importance of anomalies in the target application, as the best-performing model overall may not be the most robust choice for handling unexpected observations.

\subsection{Performance by problem characteristics}\label{sec:prob_char}

\subsubsection{Forecasting horizon}

\begin{figure}[!ht]
    \centering
    \includegraphics[width=\textwidth, trim=0cm 0cm 0cm 0cm, clip=TRUE]{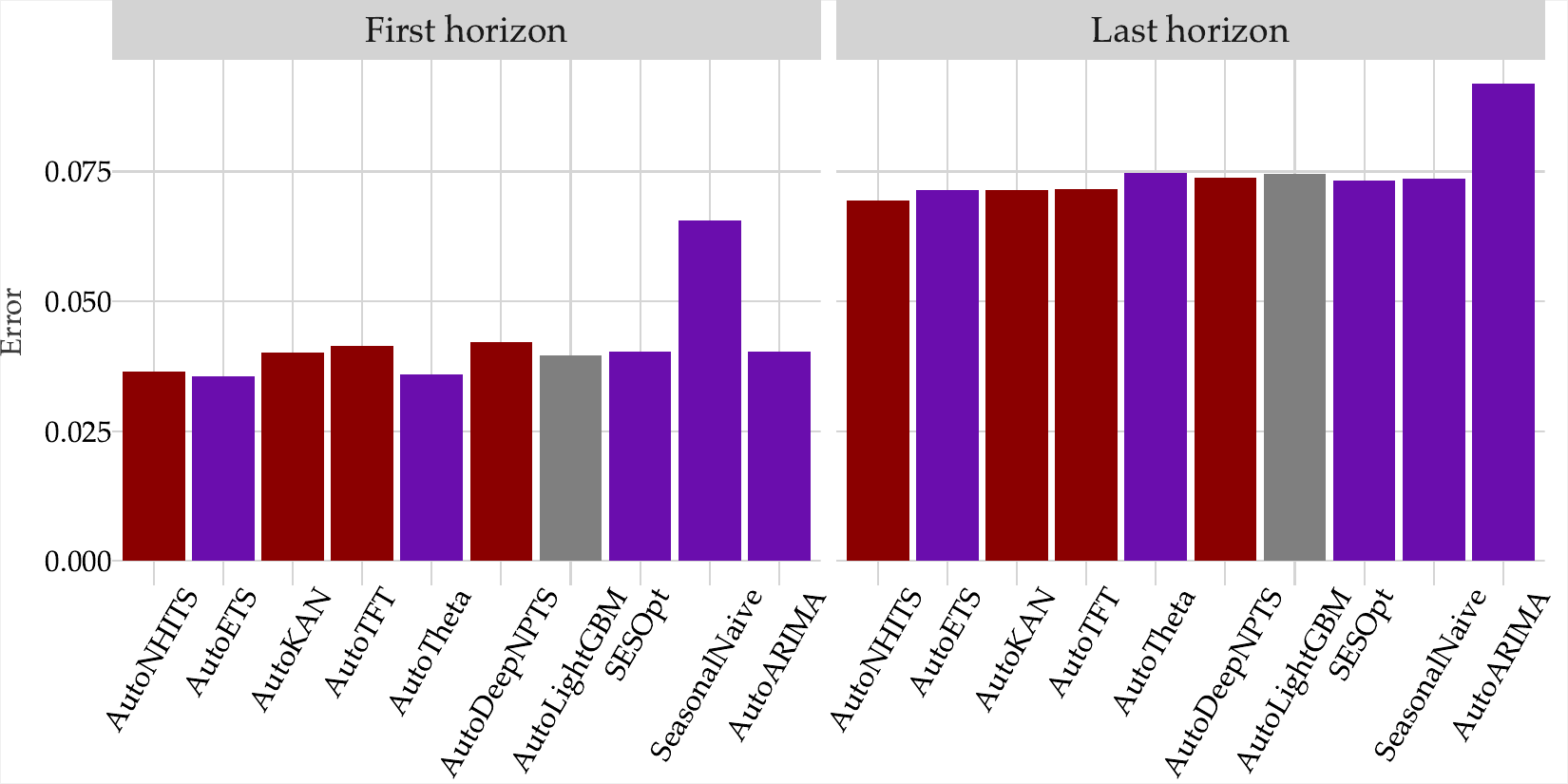}
    \caption{SMAPE scores of each model controlling for horizon condition.}
    \label{fig:plot5_hb}
\end{figure}

We also controlled the experiments for forecasting horizon. We measured performance in the first and last horizon of each series, where the former equates to one-step-ahead forecasting. The forecasting horizon varies by sampling frequency (c.f. Table \ref{tab:data}), meaning the last horizon is different for monthly (12) and quarterly (4) series.

The results (Figure \ref{fig:plot5_hb}) reveal a clear pattern in relative performance across horizons. For the first horizon, classical methods show superior performance, with \texttt{AutoETS} and \texttt{AutoTheta} outperforming all other approaches. However, this advantage diminishes for longer horizons, where neural networks become more competitive. Particularly, \texttt{AutoNHITS} shows the best performance on the last horizon, while classical methods exhibit performance degradation.
These findings align with previous research by Tang et al. \cite{tang1991time}, who also found neural networks to be more effective for long-term forecasting.

\subsubsection{Hard time series}

\begin{figure}[!ht]
    \centering
    \includegraphics[width=\textwidth, trim=0cm 0cm 0cm 0cm, clip=TRUE]{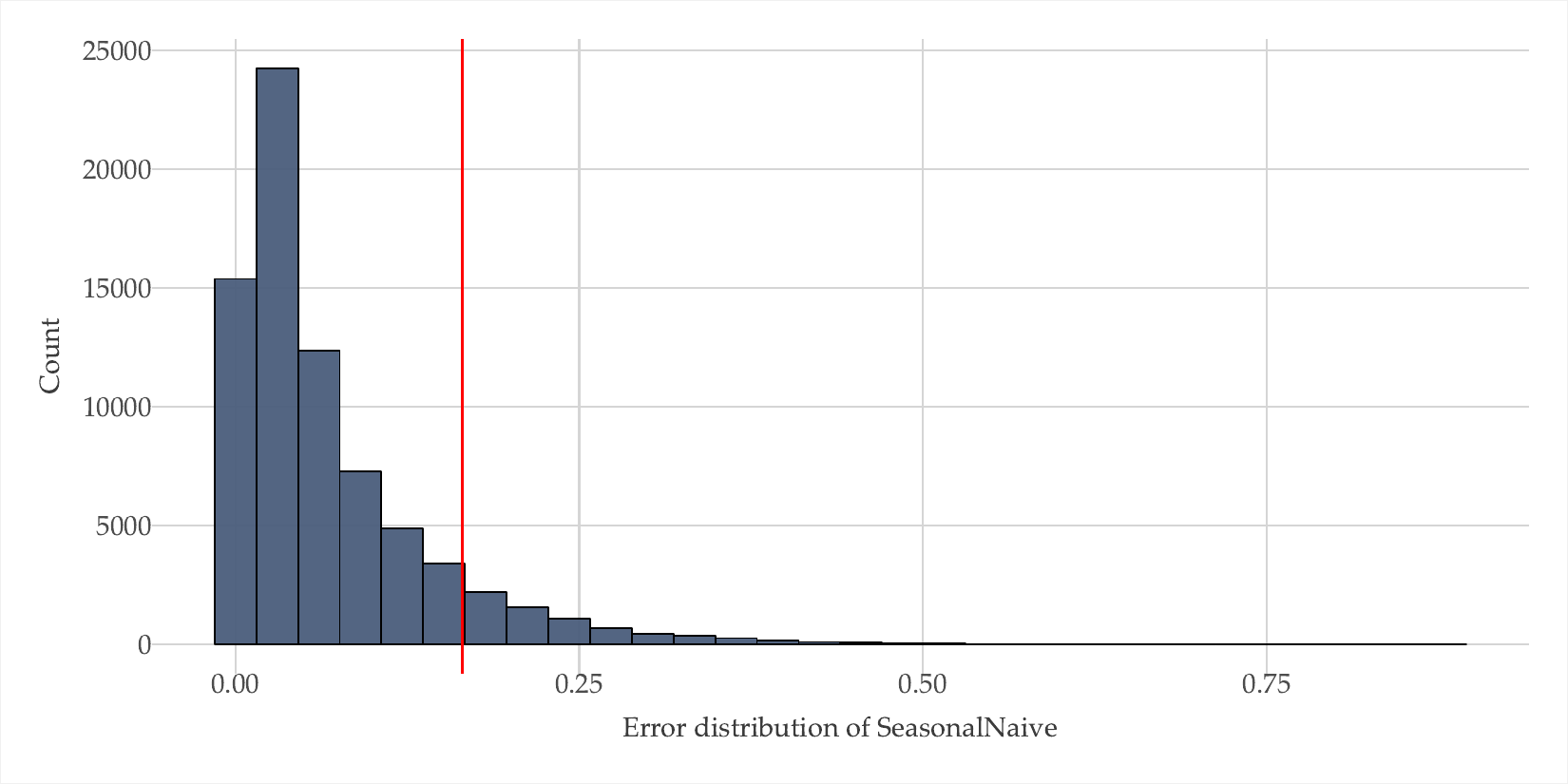}
    \caption{Distribution of SMAPE scores for \texttt{SeasonalNaive} across all time series.}
    \label{fig:plot10_baseline_dist}
\end{figure}

So far, we considered all time series in our analysis. However, some time series may exhibit patterns easily captured by a simple model. Thus, we repeat the analysis only considering hard problems. We took a data-driven and model-based approach to define a hard problem based on the performance of a baseline, namely \texttt{SeasonalNaive}.
Figure \ref{fig:plot10_baseline_dist} shows the distribution of SMAPE performance by \texttt{SeasonalNaive} across all time series. The vertical line depicts the 90\% score percentile. We consider a hard problem to be any time series corresponding to the right side of the vertical line.

\begin{figure}[!t]%
    \centering
    \subfloat[\centering Average SMAPE]{{\includegraphics[width=.48\textwidth]{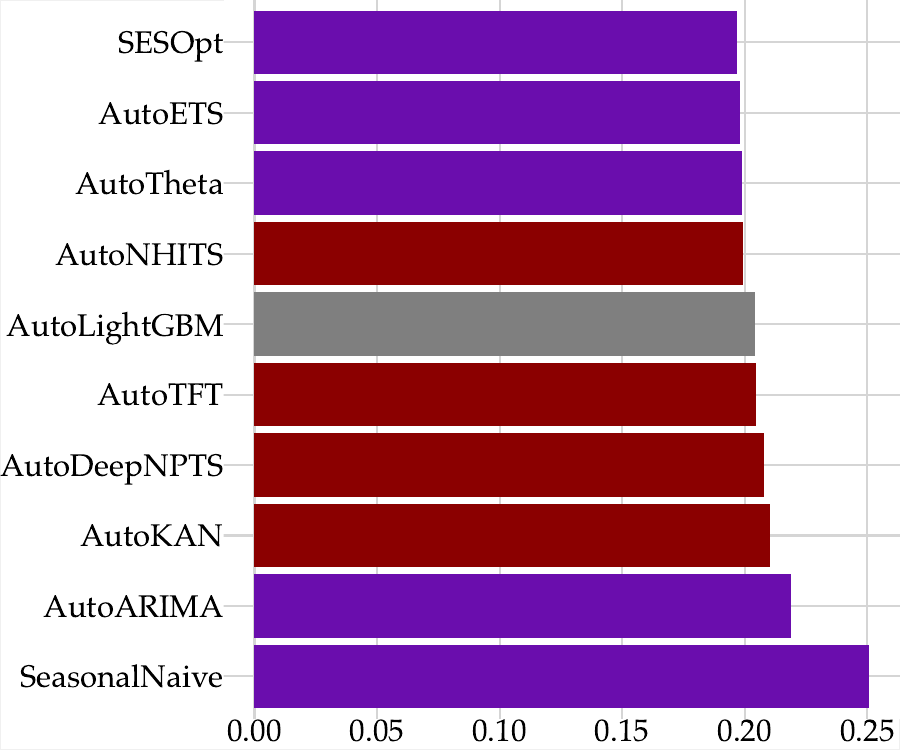} }}%
    \quad
    \subfloat[\centering SMAPE expected shortfall]{{\includegraphics[width=.48\textwidth]{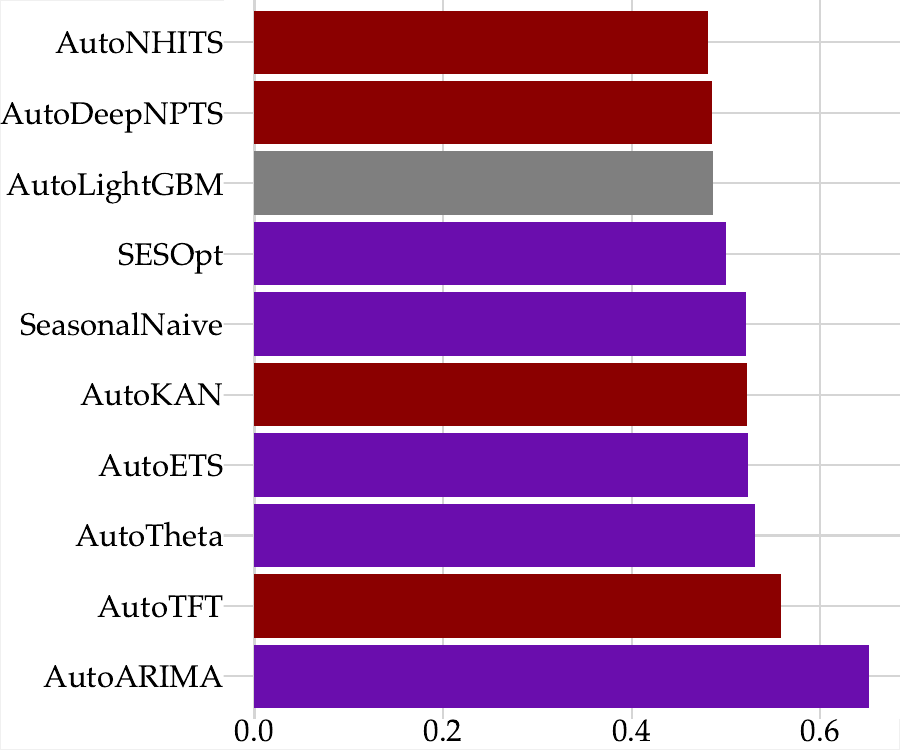}}}%
    \caption{Performance scores of each model on hard time series}%
    \label{fig:plot8_accuracy_on_hard}%
\end{figure}

Figure \ref{fig:plot8_accuracy_on_hard} presents model performance on these challenging time series. The results show a significant shift in relative performance compared to the overall analysis. Classical approaches become particularly competitive, with \texttt{SESOpt} achieving the best average performance, followed by \texttt{AutoETS} and \texttt{AutoTheta}. This suggests that simpler, well-established methods might be more robust when dealing with inherently difficult forecasting problems (as defined by \texttt{SeasonalNaive}'s performance).

While classical methods lead in average performance, \texttt{AutoNHITS} shows superior performance in the worst-case scenarios, indicating better consistency across challenging cases. Note that here, the expected shortfall score is computed using the 10\% worst cases in the subset of hard time series.
The machine learning approach \texttt{AutoLightGBM} also demonstrates improved relative performance on hard problems, though not matching the effectiveness of classical methods or \texttt{AutoNHITS}.

\section{Discussion}\label{sec:discussion}

\subsection{Practical Implications}

ModelRadar shows that evaluating forecasting models through multiple lenses provides deeper insights than traditional aggregate metrics. This multi-dimensional view enables better-informed model selection and highlights opportunities for developing more robust forecasting approaches (RQ1).

Our aspect-based evaluation framework revealed several key insights about forecasting model performance. While \texttt{AutoNHITS} emerged as the best approach across multiple dimensions, it shows limitations in specific scenarios such as one-step ahead forecasting, anomalous observations, and hard problems. Although computing time was not evaluated, \texttt{AutoNHITS} is notably more computationally efficient than most other models \cite{challu2023nhits}, especially transformer-based or recurrent-based neural networks.

We also discovered several factors that give a more nuanced perspective about the relative univariate time series forecasting accuracy (RQ2):
\begin{enumerate}
    \item Worst-case scenarios: Considering the worst-case scenarios based on expected shortfall, \texttt{AutoNHITS} demonstrates better robustness than other approaches. 
    
    \item Win/draw/loss ratios: While \texttt{AutoNHITS} shows better SMAPE scores overall, there is a reasonable chance that other approaches, including \texttt{SeasonalNaive}, outperform it, even with an equivalence margin of 10\%. This implies that the superiority of \texttt{AutoNHITS} (or any given model) is not guaranteed in all cases.
    
    \item Stationarity: Performance varies notably between stationary and non-stationary series. While \texttt{AutoNHITS} excels on non-stationary time series, \texttt{AutoTheta} emerges as the best performer on stationary ones, suggesting that different approaches may be optimal depending on stationarity properties.
    
    \item Seasonality: Methods show contrasting abilities in handling seasonality. \texttt{SESOpt} performs well on non-seasonal data but struggles with seasonal patterns. \texttt{AutoNHITS} and \texttt{AutoETS} maintain robust and competitive performance across both conditions.
    
    \item Forecasting horizon: Classical methods like \texttt{AutoETS} and \texttt{AutoTheta} excel at one-step-ahead forecasting (the first horizon step), while neural networks, particularly \texttt{AutoNHITS}, become more competitive for longer horizons. 

     \item Anomaly handling: \texttt{AutoNHITS} and other neural networks are outperformed by several classical methods when dealing with anomalous observations. This suggests that these may struggle with handling outliers or unexpected data points compared to classical forecasting techniques.
    
    \item Difficulty of problems: For challenging time series (as measured by \texttt{SeasonalNaive} performance), classical approaches become more competitive, with \texttt{SESOpt} showing the best average performance. However, \texttt{AutoNHITS} maintains superior performance in worst-case scenarios.

\end{enumerate}

Overall, our findings have several implications for practitioners. First, while \texttt{AutoNHITS} shows strong overall performance, the choice of forecasting method should be tailored to specific data characteristics and application requirements. For instance, if accurate forecasting during anomalous periods is important, classical methods such as \texttt{AutoTheta} or \texttt{AutoETS} may be preferable due to their robustness in these conditions. Moreover, classical methods based on exponential smoothing (e.g., \texttt{AutoETS}, \texttt{SESOpt}) have shown to be more competitive when dealing with hard time series forecasting problems. Similarly, for applications requiring only short-term forecasts, simpler approaches such as \texttt{AutoETS} may be more appropriate, as they have shown relatively strong performance in one-step-ahead forecasting scenarios (RQ3).

The framework also suggests that maintaining multiple models might be beneficial, as different methods excel under different conditions. This is particularly relevant for large-scale forecasting applications where time series can exhibit varying characteristics.

\subsection{Limitations}

Several limitations of our study should be noted. First, while we covered multiple aspects of forecasting performance, computational requirements were not evaluated. Second, our analysis focused on monthly and quarterly data, leaving open questions about performance on other frequencies. Finally, the definition of hard problems and anomalies could be expanded to include other criteria beyond \texttt{SeasonalNaive} performance.

Future research could extend this work in several directions. These include additional dimensions such as computational efficiency, or exploring how these findings translate to probabilistic forecasting problems.

\section{Conclusions}\label{sec:conclusions}

This paper presents an extensive empirical comparison of forecasting methods, encompassing classical approaches, machine learning algorithms, and deep learning models. Contrary to previous studies that rely on aggregate metrics, we propose ModelRadar, a framework for evaluating forecasting performance across multiple dimensions such as data characteristics and forecasting conditions.

Our analysis revealed several key findings. While \texttt{AutoNHITS} demonstrated superior overall performance according to SMAPE, its advantages varied significantly across different conditions. Classical methods proved more effective for one-step-ahead forecasting and handling anomalies, while neural approaches excelled at longer horizons. The framework also highlighted that even simpler methods such \texttt{AutoETS} and \texttt{AutoTheta} remain competitive under specific conditions, outperforming more complex approaches in a substantial percentage of cases.

These findings highlight the limitations of evaluating forecasting models through a single aggregate metric. Performance characteristics that might be crucial for specific applications - such as robustness to anomalies or accuracy at different forecasting horizons - can be masked when using traditional evaluation approaches. Our work demonstrates that a multi-dimensional evaluation framework provides practitioners with richer insights for model selection while revealing opportunities for developing more robust forecasting methods.

We believe this aspect-based evaluation approach will drive future research in several directions, from improving model robustness across different conditions to developing adaptive frameworks that leverage the complementary strengths of different forecasting methods.

\section*{Declarations}

\begin{itemize}
    \item Funding: This work was partially funded by projects AISym4Med (101095387) supported by Horizon Europe Cluster 1: Health, ConnectedHealth (n.º 46858), supported by Competitiveness and Internationalisation Operational Programme (POCI) and Lisbon Regional Operational Programme (LISBOA 2020), under the PORTUGAL 2020 Partnership Agreement, through the European Regional Development Fund (ERDF) and Agenda “Center for Responsible AI”, nr. C645008882-00000055, investment project nr. 62, financed by the Recovery and Resilience Plan (PRR) and by European Union -  NextGeneration EU, and also by FCT plurianual funding for 2020-2023 of LIACC (UIDB/00027/2020 UIDP/00027/2020). The computational resources used in this study were made available by the Portuguese National Distributed Computing Infrastructure (INCD) through the FCT Advanced Computing Projects 2024.12160.CPCA.A0.
    
    \item Conflicts of interest/Competing interests: The authors have no relevant financial or non-financial interests to disclose;
    
    \item Ethics approval: Not applicable;
    
    \item Consent to participate: Not applicable;
    
    \item Consent for publication: Not applicable;
    
    \item Availability of data and material: All experiments and data are publicly available (c.f. footnote 1);
    
    \item Code availability: Publicly available (c.f. footnote 1);
    
    \item Contributions: All authors contributed to writing and research.
\end{itemize}

\bibliographystyle{spmpsci}      

\end{document}